\pdfoutput=1

\documentclass[11pt]{article}

\usepackage{acl}
\usepackage{booktabs}
\usepackage{pbox}
\usepackage{amsmath}
\usepackage{amssymb}
\usepackage{graphicx}
\usepackage{multirow}
\usepackage{tikz}
\newcommand*\circled[1]{\tikz[baseline=(char.base)]{\node[shape=circle,draw,inner sep=0.5pt] (char) {#1};}}
\usepackage{mfirstuc}
\usepackage{makecell}
\usepackage{url}

\usepackage{times}
\usepackage{latexsym}

\usepackage[T1]{fontenc}

\usepackage[utf8]{inputenc}

\usepackage{microtype}

\title{BERT-Flow-VAE: A Weakly-supervised Model for Multi-Label Text Classification}

\author{Ziwen Liu \\
  University College London \\
  \texttt{z.liu.19@ucl.ac.uk} \\\And
  Scott Allan Orr \\
  University College London \\
  \texttt{scott.orr@ucl.ac.uk} \\
  {\bf Josep Grau-Bov\'{e}} \\
  University College London \\
  \texttt{josep.grau.bove@ucl.ac.uk} \\}

\begin{document}
\maketitle
\begin{abstract}

Multi-label Text Classification (MLTC) is the task of categorizing documents into one or more topics. Considering the large volumes of data and varying domains of such tasks, fully-supervised learning requires manually fully annotated datasets which is costly and time-consuming. In this paper, we propose BERT-Flow-VAE (BFV), a Weakly-Supervised Multi-Label Text Classification (WSMLTC) model that reduces the need for full supervision. This new model: (1) produces BERT sentence embeddings and calibrates them using a flow model, (2) generates an initial topic-document matrix by averaging results of a seeded sparse topic model and a textual entailment model that only require surface name of topics and 4-6 seed words per topic, and (3) adopts a VAE framework to reconstruct the embeddings under the guidance of the topic-document matrix. Finally, (4) it uses the means produced by the encoder model in the VAE architecture as predictions for MLTC. Experimental results on 6 multi-label datasets show that BFV can substantially outperform other baseline WSMLTC models in key metrics and achieve approximately 84\% performance of a fully-supervised model.
\end{abstract}

\section{Introduction}

As vast numbers of written comments are posted daily on social media and e-commerce platforms, there is an increasing demand for methods that efficiently and effectively extract useful information from this unstructured text data. One of the methods to analyze this unstructured text data is to classify them into organized categories. This can be considered as a Multi-label Text Classification (MLTC) task since a single data may contain multiple non-mutually-exclusive topics (aspects). There are a range of relevant applications of this task such as categorizing movies by genres \cite{hoang2018predicting}, multi-label sentiment analysis \cite{almeida2018applying} and multi-label toxicity identification \cite{gunasekara2018review}.

Fully-supervised learning methods are undesirable for this task, because of the diversity of domains of application and cost of manual labelling \cite{brody2010unsupervised}. Seeded topic models, such as SeededLDA and CorEx \cite{jagarlamudi2012incorporating,gallagher2017anchored}, where users can designate seed words as a prior to guide the models to find topics of interest, can be seen as a Weakly-Supervised Multi-Label Text Classification (WSMLTC) method. Nevertheless, as these models are mainly statistical models based on bag-of-words representation, they fail to fully exploit key sentence elements such as context and word positions. In contrast, large pre-trained language models such as BERT and GPT-3 \cite{devlin2018bert,brown2020language} produce contextualized embeddings for each word in a sentence, which has afforded them great success in the NLP field \cite{minaee2021deep,ethayarajh2019contextual}. 

Recently, prompt-based Few-Shot Learning (FSL) and Zero-Shot Learning (ZSL) methods \cite{yin2019benchmarking,yin2020universal,gao2020making} that take advantage of the general knowledge of large pre-trained language models can also approach MLTC tasks using only a few examples or topic surface names as a means of supervision. Specifically, these models convert text classification to a textual entailment task by preparing a template such as 'This example is about $\_$' as input, and then estimating the probability of the model filling the blank with certain topic names. However, this method does not work well for abstract topics and there is no agreed way to use multiple words for the entailment task.

In this paper, we propose BERT-Flow-VAE (BFV), a WSMLTC model. It is based on the Variational AutoEncoder (VAE) \cite{kingma2013auto} framework to reconstruct the sentence embeddings obtained from distil-BERT \cite{sanh2019distilbert}. Inspired by the work of \cite{li2020sentence}, we use a shallow Glow \cite{kingma2018glow} model to map the sentence embeddings to a standard Gaussian space before feeding them into the VAE model. Finally, we use the averaged results of a seeded sparse topic model and a ZSL model to guide our model to build latent variables towards pre-specified topics as predictions for MLTC.


Our contributions can be listed as follows: (1) We propose BFV, a WSMLTC model based on VAE framework, that can achieve comparable performance to a fully-supervised method on 6 multi-label datasets with only limited inputs (4 to 6 seed words per topic and surface name of topics). (2) We show that using a normalizing-flow model to calibrate sentence embeddings before feeding them into a VAE model can improve the model's MLTC performance, suggesting that pre-processing inputs is needed as it can better fit the overall objective of the VAE framework. (3) We present that the topic classification performance of ZSL method can be further improved by properly integrating predictions from a sparse seeded topic model, which complements the results from ZSL method by naturally incorporating multiple words to define a topic and could play a role of regularization.

\section{Related Work}


\paragraph{Seeded Topic Model}
Guided (seeded) topic models are built to find more desirable topics by incorporating users' prior domain knowledge. These seeded topic models can be seen as Weakly-Supervised (WS) methods to find specific topics in a corpus. \citet{andrzejewski2009latent} proposed a model by using 'z-labels' to control which words appear or not appear in certain topics. \citet{andrzejewski2009incorporating} presented DFLDA to construct Must-Link and Cannot-Link conditions between words to indirectly force the emergence of topics. \citet{jagarlamudi2012incorporating} proposed SeededLDA to incorporate seed words for each topics to guide the results found by LDA. This is achieved by biasing (1) topics to produce seed words and (2) documents containing seed words to select corresponding topics. \citet{gallagher2017anchored} presented Correlation Explanation (CorEx), a model searching for topics that are 'maximally informative'
about a set of documents. Seed words can be flexibly incorporated into the model during fitting. \citet{meng2020discriminative} proposed CatE that jointly embeds words, documents and seeded categories (topics) into a shared space. The category distinctive information is encoded during the process.

\paragraph{Weakly-supervised Text Classification}
Recently, Weakly-Supervised Text Classification (WSTC) has been rapidly developed \cite{meng2020text,wang2020x}. Most of the works used pseudo labels/documents generation and self-training. Particularly, \citet{meng2018weakly} proposed WeSTClass which uses seed information such as label surface name and keywords to generate pseudo documents and refines itself via self-training. \citet{mekala2020contextualized} proposed ConWea that uses contextualized embeddings to disambiguate user input seed words and generates pseudo labels for unlabeled documents based on these words to train a text classifier. COSINE from \cite{yu2020fine} receives weak supervision and generates pseudo labels to perform contrastive learning (with confidence reweighting) to train a classifier. Some studies integrated simple rules as weak supervision signals: \citet{ren2020denoising} used rule-annotated weak labels to denoise labels, which then supervise a classifier to predict unseen samples; \citet{karamanolakis2021self} developed ASTRA that utilizes task-specific unlabeled data, few labeled data, and domain-specific rules through a student model, a teacher model and self-training. However, these WSTC methods were specifically designed for multi-class tasks and are not optimized for WSMLTC tasks in which documents could belong to multiple classes simultaneously.

\paragraph{Prompt-based Zero-Shot Learning}
MLTC tasks can also be approached with very limited supervision by Prompt-based Few-Shot Learning (FSL) or Zero-Shot Learning (ZSL). For example, \citet{yin2019benchmarking} proposed a ZSL method for text classification tasks by treating text classification as a textual entailment problem. This model treats an input text as a premise and prepares a corresponding hypothesis (template) such as 'This example is about $\_$' for the entailment model. Finally, it uses the probability of the model filling the blank with topic names as the topic predictions. However, the choice of template and word for the entailment task requires domain knowledge and is often sub-optimal \cite{gao2020making}. Also, it is not straightforward to find multiple words as entailment for a topic. This may limit model's ability to understand abstract topics (e.g., 'evacuation' and 'infrastructure') where providing a single surface name is insufficient \cite{yin2019benchmarking}. Although some automatic search strategies \cite{gao2020making,schick2020exploiting,schick2020automatically} have been suggested, relevant research and applications are still under-explored.

\section{Proposed Model: BERT-Flow-VAE}

\subsection{Problem Formulation and Motivation}

\paragraph{Problem Formulation}

Multi-label text classification task is a broad concept, which includes many sub-fields such as eXtreme Multi-label Text Classification (XMTC), Hierarchical Multi-label Text Classification (HMTC) and multi-label topic modeling. In our model, instead of following these approaches, we follow a simpler assumption that the labels do not have a hierarchical structure and distribution of examples per label is not extremely skewed. 

More precisely, given an input corpus consisting of $N$ documents $\mathcal{D} = \{D_{1},...D_{N}\}$, the model assigns zero, single, or multiple labels to each document $D_{i} \in \mathcal{D}$ based on weak supervision signal from a dictionary of \{topic surface name:keywords\} $\mathcal{W}$ provided by user.

This is a more challenging task than multi-class text classification as samples are assumed to have non-mutually exclusive labels. This is a more practical assumption for text classification task because documents usually belong to more than one conceptual class \cite{tsoumakas2007multi}.

\paragraph{Motivation} Inspired by relevant work of VAE and $\beta$-VAE (see Appendix \ref{appendix5}), we assume that the semantic information within sentence embeddings are composed of multiple disentangled factors in the latent space. Each latent factor can be seen as a label (topic) that may appear independently. Hence, we adopted VAE as our framework to approach this task.

\subsection{Preparing the Inputs}

\paragraph{Language Model and Sentence Embedding Strategy}

As we will model the latent factors from the semantic information of sentences encoded in the word embeddings, we need to firstly convert sentences into embeddings. Specifically, given the input corpus $\mathcal{D}$, we firstly process them into a collection of sentence embeddings $E_{s} \in \mathbb{R}^{N\times V}$, where $V$ is the embedding dimension of the language model. Taking BERT as an example, there are two main ways to produce such sentence embeddings: (1) using the special token ($[CLS]$ in BERT) and (2) using a mean-pooling strategy to aggregate all word embeddings into a single sentence embedding.  We tested and showed the performance of the two versions in section \ref{sec:res}. Lastly, for computational efficiency, we used distil-BERT \cite{sanh2019distilbert} as our language model, which is a lighter version of BERT with comparable performance.

Moreover, instead of simply averaging the embeddings of words in a sentence with equal weights, we also tested a TF-IDF averaging strategy. Specifically, we firstly calculated the weights of words in a sentence using the TF-IDF algorithm with $L_2$ normalization, and then averaged the words according to the TF-IDF weights. To avoid weights of some common words to be nearly zero, we combined 10\% mean pooling weights and 90\% TF-IDF pooling weights as the final embeddings. 

\paragraph{Flow-calibration}

Sentence embeddings obtained from BERT without extra fine-tuning have been found to poorly capture the semantic meaning of sentences. This is reflected by the performance of BERT on sentence-level tasks such as predicting Semantic Textual Similarity (STS) \cite{reimers2019sentence}. This may be caused by anisotropy (embeddings occupy a
narrow cone in the vector space), a common problem of embeddings produced by language models \cite{ethayarajh2019contextual,li2020sentence}. To address this problem, following the work of \cite{li2020sentence}, we adopted BERT-Flow to calibrate the sentence embeddings. More exactly, we used a shallow Glow \cite{kingma2018glow} with K = 16 and L = 1, a normalizing-flow based model, with random permutation and affine coupling to post-process the sentence embeddings from all 7 layers of distil-BERT (including the word embedding layer). We tested different combinations of the 7 post-processed embeddings and took the average of embeddings from the first, second and sixth layer based on the metrics evaluated on the STS benchmark dataset.


Since normalizing-flow based models can create an invertible mapping from the BERT embedding space to a standard Gaussian latent space \cite{li2020sentence}, the advantages of using flow calibration are: (1) it improves the anisotropy to make the sentence embeddings more semantically distinguishable, and (2) it converts the distribution of BERT embeddings to be standard Gaussian, which fits the objective of minimizing mean-squared reconstruction error and Kullback–Leibler Divergence (KLD) with a standard Gaussian prior distribution in the following VAE model. 

\paragraph{Backend Model}

To guide our model towards some pre-specified topics, we used Zero-Shot Text Classification method (0SHOT-TC) proposed by \cite{yin2019benchmarking} as the backend model. Specifically, we used RoBERTa-large \cite{liu2019roberta} as the language model for 0SHOT-TC. Following the example mentioned previously, we prepared a template (hypothesis) with the shape 'This example is about $\_$' for each sentence (premise) and filled the blank with the surface name of topics. Finally, we took the probability of \emph{entailment} as that of the topic appearing in the sentence for each class and collected this as $\mathcal{T}_{0SHOT-TC} \in \mathbb{R}^{N\times M}$, where $M$ is the number of topics.

However, because current zero-shot learning methods lack an agreed way to find multiple words as entailment for a topic, we further used a seeded topic model as a complement. More exactly, we selected Anchored Correlation Explanation (CorEx) \cite{gallagher2017anchored} as another backend model. By following the approach used by \cite{jagarlamudi2012incorporating,gallagher2017anchored}, we randomly chose 4 to 6 seed words from the top 20 most discriminating words of each topic as seed words to better simulate real-world applications. Finally, we estimated unnormalized document-topic matrix $\mathcal{T}_{CorEx} \in \mathbb{R}^{N\times M}$ and took the combination:
\begin{align*}
\mathcal{T} = \omega\times\mathcal{T}_{0SHOT-TC}+(1-\omega)\times\mathcal{T}_{CorEx}
\end{align*}
\noindent where $\omega$ is the combination weight. We set $\omega$ = 0.5 herein (details will be discussed in section \ref{as}). 

\subsection{Model Description}

\paragraph{Model Architecture and Objective Function}

\begin{figure}
\includegraphics[width=\linewidth]{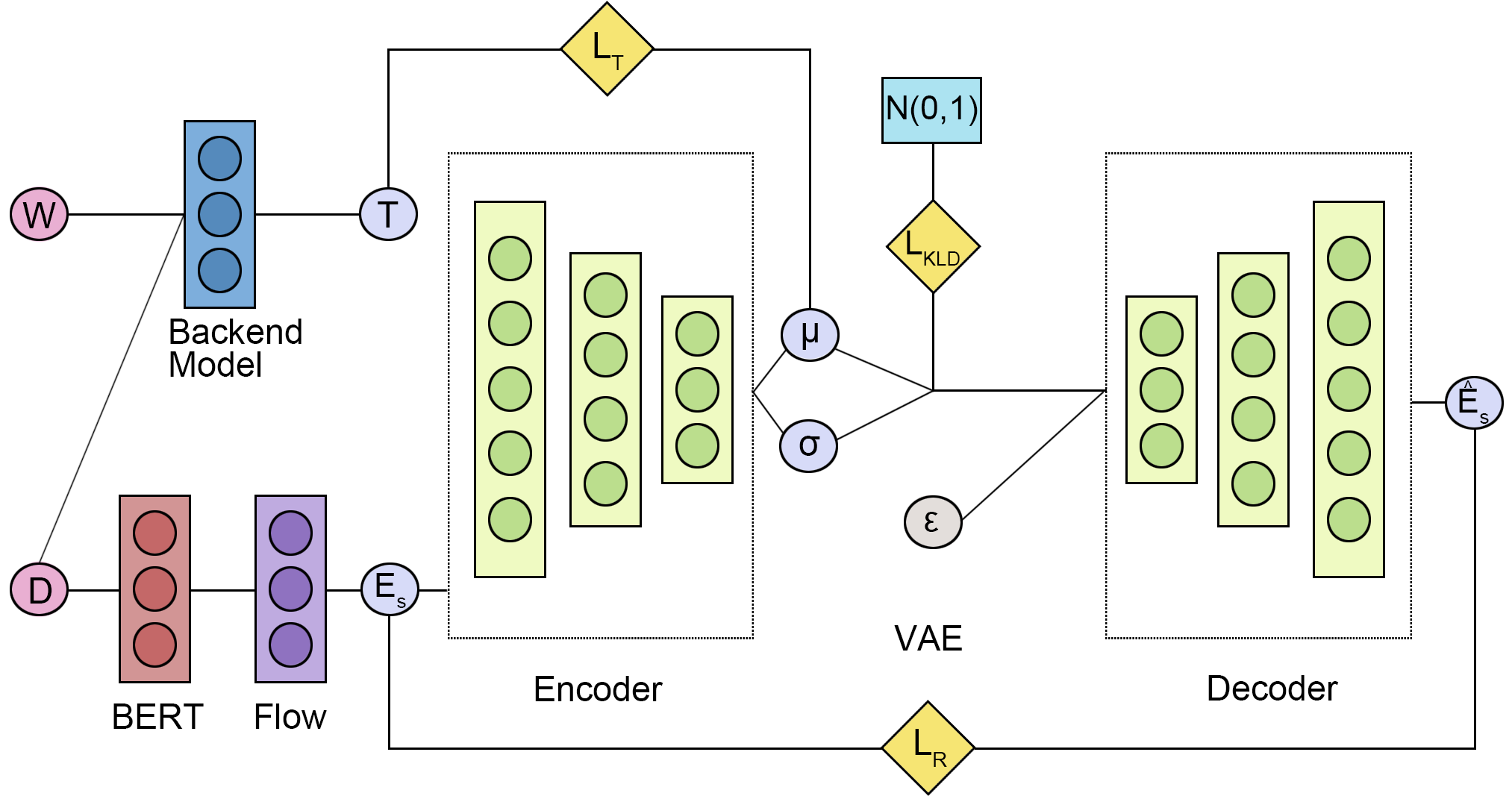}
\caption{Architecture of the proposed model. 
}
\label{model_archi}
\end{figure}

An overview of the model architecture can be seen in Fig \ref{model_archi}. Specifically, we used fully connected layers combined with layer normalization \cite{ba2016layer} and Parametric ReLU (PReLU) \cite{he2015delving}. The encoder model $q_{\phi}$ receives flow-calibrated sentence embeddings $E_{s}$ and outputs mean ($\mu \in \mathbb{R}^{N\times M}$) and variance ($\sigma \in \mathbb{R}^{N\times M}$) which will be the inputs to the decoder model $p_{\theta}$ to produce reconstructed sentence embeddings $\hat{E}_{s} \in \mathbb{R}^{N\times V}$. 

As in the vanilla VAE model in Appendix \ref{appendix5}, the objective function of our model contains a reconstruction loss and KLD loss. We used the mean-squared error between $E_{s}$ and $\hat{E}_{s}$ as the reconstruction loss because input embeddings have been calibrated to have a standard Gaussian distribution, and used the KL divergence between the output ($\mu$ and $\sigma$) of the encoder and the prior $\mathcal{N}(0,I)$ as the KLD loss. In addition, in order to guide the model's direction towards the pre-specified topics, we added another loss term dubbed topic loss: 


\begin{align*}
L_{\mathcal{T}} &=
-\frac{1}{NM}\sum^{N}_{i}\sum^{M}_{j}\mathcal{T}_{ij} \cdot log(sigmoid(\mu_{ij}))
\end{align*} 

\noindent where $sigmoid(\cdot)$ is the element-wise sigmoid function. $L_{\mathcal{T}}$ is the binary cross-entropy between $\mu$ and $\mathcal{T}$ to encourage $\mu$ to be closer to $\mathcal{T}$.

Notice that the value of $sigmoid(\mu) \in \mathbb{R}^{N\times M}$ produced by the encoder can be viewed as a document-topic matrix. Thus, we used it as the model's prediction for MLTC. $sigmoid(\mu_{ij}) > 0.5$ is be predicted as positive (i.e., topic $j$ appears in $i$th document). $\sigma$ is be left as free values to reconstruct $\hat{E}_{s}$. 


As shown by \cite{higgins2016beta,burgess2018understanding}, the weight of different components in the objective function of the VAE model is important to find disentangled representations. In particular, based on our observations, the ratio between $L_{KLD}$ and $L_{\mathcal{T}}$ is be crucial. Hence, we set a hyper-parameter $\gamma$ in the objective function controlling the ratio of $L_{KLD}$ and $L_{\mathcal{T}}$. Finally, the objective function of our VAE model is:

\begin{align*}
\mathcal{L}(\theta,\phi;E_{s},\mathcal{T},\alpha,\eta) = -(L_{R} + \alpha L_{KLD} + \eta L_{\mathcal{T}}) 
\end{align*}

\noindent where $\alpha = 0.1\times\sqrt{\gamma}$ and $\eta = 0.1\times\gamma M$. It can be seen that a higher value of $\gamma$ will lead to a heavier penalty on $L_{\mathcal{T}}$, and therefore $\mu$ will become more similar to $\mathcal{T}$; while, conversely, a lower value of $\gamma$ will make $\mu$ diverge from $\mathcal{T}$. As $L_{KLD}$ pushes $sigmoid(\mu)$ towards 0.5, a lower value of $\gamma$ would encourage model to make aggressive predictions.

\paragraph{Hyper-Parameter Scheduling (HPS)}

We adopted the strategy to gradually change the hyper-parameters of the model. We used the first epoch of training as a warm-up stage \cite{sonderby2016ladder} by setting the value of $\alpha$ to one-tenth of its original value. Similarly, we also halved the value of $\eta$ in the last epoch of training, aiming to reduce the dependency of $\mu$ on $\mathcal{T}$.

\section{Experiments Setup}

\subsection{Datasets}

We chose \emph{Restaurant} and \emph{Laptop} datasets from SemEval 2014 Task 4 \cite{pontiki-etal-2014-semeval} as well as \emph{CitySearch} \cite{ganu2009beyond}, which are very popular in aspect-based sentiment analysis studies. We also added \emph{SentiHood} \cite{saeidi-etal-2016-sentihood}, a similar dataset to the SemEval datasets. Additionally, \emph{Reuters} (ApteMod version) \cite{APTE94}, a well-known text classification dataset containing Reuters financial newswire service in 1987 was also selected. Finally, given that the Restaurants, CitySearch and SentiHood are all largely related to food-related aspects, we added a self-made dataset called \emph{Heritage}, which is composed of 3,760 online reviews for 77 heritage sites. There are 9 categories in this dataset: \emph{heritage, exhibition, price, family, service, transport, facilities, environment} and \emph{miscellaneous} (see Appendix \ref{appendix3}).

\subsection{Data Pre-processing}

\paragraph{Pre-processing}

For all datasets, we only kept categories which have at least 30 corresponding samples or account for at least 1\% in the whole dataset. Also, we removed categories with no specific meaning (e.g., 'miscellaneous'). Finally, for datasets which do not have a pre-specified train-test split, we used 20\% of the examples as the test set (splits were carefully conducted to account for class balance). Table \ref{dataset_summary} shows detailed information on the datasets after pre-processing.


\begin{table*}
\resizebox{\textwidth}{!}{
\begin{tabular}{lcccccc}
\toprule
{Dataset} & {Number of samples} & {Number of classes} & {\makecell{Max number of data\\ per class}} & {\makecell{Min number of data\\ per class}} & {\makecell{Multilabel \\ ratio}} & {\makecell{Average\\text length}} \\
\midrule
Reuters & 10788 & 42 & 3964 & 31 & 0.14 & 128 \\
CitySearch & 3315 & 4 & 1227 & 298 & 0.11 & 14 \\
Sentihood & 5215 & 9 & 714 & 143 & 0.10 & 15 \\
Restaurant & 3844 & 4 & 1651 & 402 & 0.16 & 13 \\
Laptop & 3308 & 5 & 639 & 175 & 0.12 & 13 \\
Heritage & 3760 & 8 & 623 & 156 & 0.11 & 18 \\
\bottomrule
\end{tabular}
}
\caption{Metadata for 6 selected datasets after pre-processing. "Multilabel ratio" is the percentage of samples having more than one label.}
\label{dataset_summary}
\end{table*}

\paragraph{Model Fine-tuning and Training}


We finetuned distil-BERT before using it to produce embeddings. Specifically, we used a learning rate of $1\times 10^{-5}$ for the word embedding layer and top 3 layers of distil-BERT, and $1\times 10^{-3}$ for last 3 layers. The weight decay was set to 0 for the bias and layer normalization weights. We used a warm-up strategy to gradually unfreeze the parameters of distil-BERT. We ran the fine-tuning process for 5 epochs on each dataset separately. Additionally, we trained the Glow model with a learning rate of $1\times 10^{-3}$ for 5 epochs. Lastly, we trained the BFV model with a learning rate of $1\times 10^{-3}$ for 10 epochs, with weight decay set using the same values as the distil-BERT fine-tuning. AdamW \cite{loshchilov2017decoupled} is the optimizer used for training all three models.

\subsection{Evaluation Procedure}

\paragraph{Evaluation Metrics}

One of the limitations of this model is that, as we only reconstruct the pooled representation of the sentences rather than words, there is no explicit modelling of word-topic relationship. Therefore, metrics for topic models such as perplexity and topic coherence cannot be directly measured. Thus, they will not be reported. We calculated the macro-average of each class for the metrics defined only for binary predictions such as F1-score. More detailed definition of metrics can be found in Appendix \ref{appendix4}. We report the average and standard deviation over 10 runs for our model and its ablated versions.

\paragraph{Baseline Models}
In this paper, we only compare our model with methods that can perform multi-label prediction with weak supervision (only surface names and keywords are provided). To the best of the authors' knowledge, there are few methods built specifically for the WSMLTC tasks. Therefore, we consider guided topic models and prompting based zero-shot learning language models as the work most closely related to our model as baselines. For weakly-supervised methods that were specifically designed for multi-class tasks, we compare our model with them in Appendix \ref{appendix2} since we converted them to perform MLTC tasks which may cause them to perform differently from their original multi-class design.

We used backend models (\textbf{CorEx}, \textbf{0SHOT-TC} and \textbf{CorEx+0SHOT-TC}) and \textbf{Guided LDA (GLDA)}, a standard LDA with initial word-topic priors biased towards seeded topics, as baseline models to compare the performance of our models. Following the work in \cite{brody2010unsupervised}, we used a POS-tagging model\footnote{\url{huggingface.co/vblagoje/bert-english-uncased-finetuned-pos}} to only keep nouns and adjectives for GLDA and CorEx. We also included ablated versions of BFV:

\textbf{BERT-Whitening-VAE (BWV)}: The Flow in BFV is replaced by a simple whitening operation introduced by \cite{huang-etal-2021-whiteningbert-easy}.

\textbf{BERT-VAE (BV)}: No calibration is performed to the BERT-embeddings.

\textbf{BERT-Flow-VAE-CLS (BFVCLS)}: Instead of using pooling average strategy, $[CLS]$ token is used as the representation of a sentence.

\textbf{BERT-Flow-Encoder (BFE)}: Only the encoder part of the VAE is kept, and two loss components ($L_{R}$ and $L_{KLD}$) are disabled.

\textbf{BERT-Flow (BF)}: The VAE part in BFV is replaced by a textual entailment classification header \footnote{\url{huggingface.co/huggingface/distilbert-base-uncased-finetuned-mnli}} with the same template. The classification header has been finetuned on the MNLI dataset to adjust for the flow-calibrated model.

\textbf{0SHOT-TC-Flow}: Flow calibration is applied to the original 0SHOT-TC model. The classification header has been finetuned on the MNLI dataset to adjust for the flow-calibrated model.

Lastly, to show the performance gap between our model and fully-supervised methods, the performance of the \textbf{BFE} trained in a fully-supervised way using groundtruth labels (\textbf{BFE-Sup}) is included.

\section{Results}
\label{sec:res}
\subsection{Quantitative Evaluation}

Table \ref{classification} presents the quantitative evaluation of the models measured using classification metrics (clustering performance and qualitative evaluation can be found in Appendix \ref{appendix1}.). We make the following observations based on these results: 

(1) BFV outperforms other WSMLTC models in most datasets with relatively large margin except for \emph{Reuters}, in terms of F1-score and APS.

(2) In terms of the F1-score, BFV with only seed words and surface name of topics as supervision can, on average, achieve approximately 84\% performance of the fully-supervised model as reflected in Table \ref{f1p}. In addition, the performance gap between weakly-supervised and fully-supervised model is narrow in datasets of social media reviews. This suggests that the weakly-supervised models are able to find topics with limited guidance in the context of short text, informal and polysemous words that characterise reviews.

\begin{table}
\renewcommand\thetable{3}
\begin{tabular}{lc}
\toprule
 & \% of BFE-Sup's F1 \\
\midrule
BFE-Sup & 100.0 (100.0) \\
BFV & 84.38 (83.40) \\
BWV & 78.67 (77.49) \\
BV & 73.47 (71.28) \\
BFVCLS & 58.93 (56.10) \\
BFE & 69.15 (67.46) \\
BF & 26.57 (25.50) \\
0SHOT-TC+CorEx & 57.84 (53.97) \\
0SHOT-TC-Flow & 33.42 (33.56) \\
0SHOT-TC & 69.57 (71.53) \\
CorEX & 53.86 (50.25) \\
GLDA & 27.84 (27.37) \\
\bottomrule
\end{tabular}
\caption{Models performance in percentage of the fully-supervised model measured by F1-score averaged across all datasets. Numbers in the brackets are F1-score averaged across 5 datasets excluding \emph{Heritage}.}
\label{f1p}
\end{table}

(3) After replacing Flow with a simple whitening operation, there is no significant drop of performance of the model in terms of key metrics such as F1, demonstrating the robustness of the model.

\begin{table*}[htbp]
\renewcommand\thetable{2}
\centering
\resizebox{0.85\textwidth}{!}{
\begin{tabular}{llllllllll}
\toprule
{} & {} & {ACC} & {HS} & {P@3} & {\textbf{F1}} & {Recall} & {Precision} & {\textbf{APS}} & {\textbf{AUC}} \\
\midrule
\multirow[c]{13}{*}{Reuters} & BFE-Sup (20) & 75.52 & 80.82 & 82.86 & 63.29 & 54.12 & 83.00 & 70.18 & 97.91 \\
 & BFV (20) & 44.65 (1.33) & 55.74 (1.35) & 66.03 (1.5) & \textbf{44.99} (0.51) & 55.38 (0.6) & \textbf{51.58} (0.84) & 56.18 (0.34) & \textbf{95.86} (0.14)\\
 & BWV (20) & 39.83 (1.23) & 51.33 (1.23) & 61.98 (1.2) & 44.59 (0.41) & 54.96 (0.5) & 50.27 (0.76) & 55.96 (0.36) & 95.85 (0.1)\\
 & BV (20) & 48.84 (2.5) & \textbf{57.72} (2.61) & 65.40 (2.88) & 40.91 (0.73) & 45.63 (1.65) & 50.56 (1.71) & 51.12 (0.7) & 95.58 (0.11)\\
 & BFVCLS (20) & \textbf{49.60} (4.92) & 56.85 (5.15) & 62.53 (5.51) & 33.97 (1.84) & 37.08 (3.32) & 44.19 (1.57) & 46.23 (0.54) & 94.37 (0.11)\\
 & BFE (20) & 45.09 (0.74) & 53.54 (0.78) & 60.37 (0.75) & 37.56 (0.36) & 39.81 (0.35) & 50.13 (1.1) & 44.21 (0.39) & 94.07 (0.09)\\
 & BF & 1.82 & 3.35 & 7.96 & 3.92 & 33.51 & 2.72 & 3.12 & 43.86 \\
 & 0SHOT-TC+CorEx & 12.62 & 25.61 & 49.30 & 25.28 & 60.56 & 21.43 & 50.28 & 93.91 \\
 & 0SHOT-TC-Flow & 0.43 & 4.76 & 11.39 & 5.22 & 58.01 & 3.31 & 3.81 & 54.94 \\
 & 0SHOT-TC & 19.68 & 43.19 & \textbf{75.26} & 44.55 & \textbf{70.65} & 38.21 & \textbf{56.22} & 91.64 \\
 & CorEx & 10.53 & 22.84 & 31.56 & 24.27 & 57.53 & 20.71 & 17.90 & 86.71 \\
 & GLDA & 7.92 & 8.28 & 8.28 & 6.88 & 4.15 & 34.33 & 26.33 & 75.18 \\
 &  &  &  &  &  &  &  &  &  \\
\multirow[c]{13}{*}{CitySearch} & BFE-Sup (1) & 71.79 & 75.72 & 76.66 & 72.66 & 64.81 & 83.13 & 82.30 & 94.37 \\
 & BFV (1) & \textbf{61.07} (1.25) & \textbf{68.95} (0.93) & \textbf{72.79} (0.75) & \textbf{69.48} (0.96) & 76.50 (1.33) & 64.92 (1.35) & \textbf{77.25} (0.42) & \textbf{92.37} (0.22)\\
 & BWV (1) & 54.60 (1.38) & 61.07 (1.39) & 62.87 (1.51) & 62.63 (1.53) & 60.99 (2.92) & 66.13 (1.92) & 69.98 (0.69) & 89.19 (0.26)\\
 & BV (1) & 53.74 (0.99) & 58.51 (1.12) & 59.34 (1.25) & 59.64 (2.2) & 52.24 (5.07) & 75.06 (3.06) & 72.41 (0.47) & 89.09 (0.25)\\
 & BFVCLS (1) & 54.39 (1.34) & 58.86 (1.37) & 59.08 (1.38) & 48.82 (1.61) & 39.67 (2.3) & \textbf{79.56} (2.76) & 67.09 (0.4) & 87.97 (0.2)\\
 & BFE (1) & 53.26 (0.68) & 57.90 (0.79) & 58.33 (0.79) & 54.28 (1.25) & 43.84 (1.09) & 78.08 (1.62) & 69.89 (0.84) & 88.63 (0.46)\\
 & BF & 29.41 & 33.74 & 33.81 & 18.80 & 21.09 & 19.04 & 19.39 & 48.41 \\
 & 0SHOT-TC+CorEx & 47.36 & 51.43 & 51.92 & 39.98 & 30.83 & 58.42 & 63.75 & 87.80 \\
 & 0SHOT-TC-Flow & 37.86 & 46.10 & 50.35 & 41.68 & 51.13 & 40.97 & 45.08 & 77.08 \\
 & 0SHOT-TC & 41.78 & 53.12 & 61.29 & 60.43 & \textbf{86.90} & 48.30 & 71.04 & 88.12 \\
 & CorEx & 45.70 & 49.57 & 50.04 & 35.66 & 26.44 & 55.57 & 43.89 & 69.97 \\
 & GLDA & 33.18 & 35.97 & 35.97 & 27.42 & 23.55 & 35.92 & 30.63 & 65.73 \\
 &  &  &  &  &  &  &  &  &  \\
\multirow[c]{13}{*}{Sentihood} & BFE-Sup (1) & 74.04 & 77.41 & 78.14 & 68.88 & 61.32 & 79.05 & 76.60 & 96.81 \\
 & BFV (1) & 42.94 (1.57) & 51.23 (1.37) & 56.35 (1.25) & \textbf{53.54} (0.88) & 75.58 (1.13) & 47.21 (0.96) & 59.10 (0.68) & \textbf{92.12} (0.14)\\
 & BWV (1) & 50.29 (1.73) & 55.61 (1.51) & 57.80 (1.36) & 51.40 (1.36) & 60.18 (2.32) & 53.49 (0.71) & 59.80 (0.33) & 91.96 (0.13)\\
 & BV (1) & 51.41 (3.61) & 56.66 (3.14) & \textbf{59.00} (2.67) & 51.02 (1.66) & 53.77 (2.94) & \textbf{59.25} (1.04) & \textbf{60.78} (0.48) & 91.93 (0.22)\\
 & BFVCLS (1) & \textbf{53.62} (2.56) & 56.13 (2.42) & 56.75 (2.17) & 32.92 (1.28) & 31.04 (2.14) & 55.77 (2.0) & 48.80 (0.58) & 89.20 (0.14)\\
 & BFE (1) & 52.60 (0.45) & \textbf{56.88} (0.4) & 58.63 (0.44) & 46.55 (0.87) & 46.83 (0.79) & 56.58 (1.09) & 56.12 (0.44) & 90.46 (0.21)\\
 & BF & 20.66 & 24.32 & 25.57 & 10.81 & 46.58 & 6.42 & 6.76 & 51.86 \\
 & 0SHOT-TC+CorEx & 53.05 & 55.90 & 56.42 & 39.17 & 39.50 & 45.05 & 53.26 & 91.29 \\
 & 0SHOT-TC-Flow & 0.60 & 6.82 & 10.02 & 11.37 & 64.15 & 6.42 & 9.03 & 48.18 \\
 & 0SHOT-TC & 4.16 & 18.33 & 36.82 & 39.13 & \textbf{89.11} & 28.16 & 55.30 & 91.12 \\
 & CorEx & 52.52 & 54.98 & 55.34 & 34.71 & 33.82 & 40.76 & 33.00 & 75.87 \\
 & GLDA & 39.17 & 40.39 & 40.40 & 16.41 & 15.43 & 20.58 & 16.70 & 71.18 \\
 &  &  &  &  &  &  &  &  &  \\
\multirow[c]{13}{*}{Restaurant} & BFE-Sup (1) & 75.38 & 80.93 & 81.72 & 81.59 & 75.57 & 89.11 & 90.02 & 95.87 \\
 & BFV (1) & \textbf{68.85} (1.95) & \textbf{74.42} (2.02) & \textbf{76.34} (2.16) & \textbf{80.49} (0.77) & 80.58 (1.72) & 80.88 (2.24) & \textbf{89.73} (0.4) & \textbf{95.21} (0.27)\\
 & BWV (1) & 58.09 (1.91) & 65.65 (1.76) & 68.13 (1.97) & 73.44 (1.35) & 80.02 (3.47) & 69.62 (3.21) & 85.55 (0.35) & 93.34 (0.17)\\
 & BV (1) & 53.66 (2.18) & 59.43 (2.36) & 60.27 (2.63) & 69.52 (1.97) & 60.85 (4.55) & \textbf{84.74} (3.94) & 84.97 (0.45) & 91.78 (0.33)\\
 & BFVCLS (1) & 57.84 (2.94) & 64.95 (2.83) & 66.02 (2.61) & 60.14 (2.89) & 53.00 (2.45) & 83.62 (8.77) & 78.28 (1.11) & 90.09 (0.34)\\
 & BFE (1) & 57.36 (1.01) & 63.38 (0.9) & 64.11 (0.86) & 69.05 (1.45) & 60.35 (1.42) & 83.59 (1.18) & 81.57 (0.59) & 91.21 (0.24)\\
 & BF & 8.38 & 21.33 & 24.74 & 30.21 & 51.64 & 25.00 & 25.62 & 49.18 \\
 & 0SHOT-TC+CorEx & 46.00 & 51.12 & 51.54 & 57.02 & 45.92 & 78.68 & 83.06 & 93.40 \\
 & 0SHOT-TC-Flow & 4.88 & 28.45 & 47.24 & 38.44 & 88.02 & 26.49 & 30.58 & 60.69 \\
 & 0SHOT-TC & 37.25 & 56.46 & 72.90 & 63.04 & \textbf{95.00} & 48.61 & 83.65 & 92.43 \\
 & CorEx & 45.25 & 50.19 & 50.31 & 54.42 & 43.00 & 77.55 & 61.20 & 74.97 \\
 & GLDA & 26.12 & 31.08 & 31.08 & 34.46 & 30.93 & 46.52 & 42.51 & 72.39 \\
 &  &  &  &  &  &  &  &  &  \\
\multirow[c]{13}{*}{Laptop} & BFE-Sup (2) & 60.15 & 63.48 & 63.90 & 49.24 & 39.87 & 65.76 & 57.66 & 87.21 \\
 & BFV (2) & 45.72 (1.18) & 49.37 (1.18) & 50.53 (1.2) & \textbf{36.40} (1.2) & 38.59 (1.54) & 42.75 (1.67) & \textbf{39.25} (0.54) & \textbf{73.38} (0.49)\\
 & BWV (2) & 45.04 (0.57) & 47.96 (0.37) & 48.76 (0.42) & 32.58 (1.15) & 32.34 (1.46) & 44.62 (0.84) & 37.56 (0.31) & 72.72 (0.18)\\
 & BV (2) & 51.18 (2.11) & 52.82 (1.86) & 53.16 (1.7) & 24.82 (1.71) & 21.52 (2.59) & 52.31 (2.46) & 37.21 (0.49) & 73.28 (0.33)\\
 & BFVCLS (2) & \textbf{52.46} (0.97) & \textbf{53.59} (0.86) & \textbf{53.79} (0.86) & 18.78 (1.56) & 14.36 (1.04) & \textbf{54.01} (3.23) & 35.23 (0.65) & 72.30 (0.36)\\
 & BFE (2) & 49.15 (0.63) & 50.96 (0.6) & 51.30 (0.55) & 25.14 (0.5) & 22.35 (0.74) & 52.14 (3.68) & 35.96 (0.69) & 71.90 (0.36)\\
 & BF & 9.90 & 18.32 & 21.60 & 21.03 & \textbf{68.05} & 12.85 & 14.60 & 54.14 \\
 & 0SHOT-TC+CorEx & 48.76 & 50.51 & 50.56 & 23.71 & 22.39 & 35.82 & 33.91 & 71.89 \\
 & 0SHOT-TC-Flow & 13.00 & 20.07 & 23.24 & 18.99 & 44.92 & 12.97 & 13.59 & 53.47 \\
 & 0SHOT-TC & 14.98 & 25.13 & 31.76 & 34.49 & 64.18 & 27.10 & 34.78 & 70.96 \\
 & CorEx & 48.39 & 50.09 & 50.12 & 23.00 & 21.89 & 34.69 & 23.64 & 60.20 \\
 & GLDA & 43.19 & 43.79 & 43.79 & 10.91 & 9.73 & 17.03 & 16.56 & 60.99 \\
 &  &  &  &  &  &  &  &  &  \\
\multirow[c]{13}{*}{Heritage} & BFE-Sup (5) & 53.44 & 56.59 & 56.81 & 49.52 & 39.42 & 67.95 & 58.86 & 89.85 \\
 & BFV (5) & 40.10 (1.41) & 47.23 (1.36) & \textbf{51.42} (1.5) & \textbf{44.20} (1.22) & 50.34 (2.11) & 41.75 (0.85) & \textbf{46.68} (0.56) & \textbf{81.55} (0.32)\\
 & BWV (5) & 33.75 (0.98) & 42.22 (0.77) & 47.29 (0.85) & 41.89 (0.64) & 51.99 (2.07) & 37.30 (0.95) & 45.08 (0.49) & 79.90 (0.21)\\
 & BV (5) & \textbf{42.37} (1.01) & \textbf{47.77} (1.11) & 50.26 (1.4) & 41.81 (1.49) & 41.18 (2.85) & 46.84 (1.39) & 45.33 (0.25) & 81.28 (0.25)\\
 & BFVCLS (5) & 41.58 (1.61) & 46.34 (1.26) & 48.61 (1.17) & 36.18 (1.25) & 33.39 (1.94) & \textbf{47.93} (3.81) & 40.75 (0.46) & 80.81 (0.15)\\
 & BFE (5) & 40.05 (1.04) & 45.65 (0.73) & 48.60 (0.59) & 38.41 (0.72) & 37.44 (0.88) & 42.86 (0.74) & 40.15 (0.46) & 80.86 (0.34)\\
 & BF & 4.41 & 11.06 & 13.44 & 15.81 & 67.63 & 9.22 & 9.65 & 49.20 \\
 & 0SHOT-TC+CorEx & 41.50 & 44.95 & 46.10 & 38.21 & 36.36 & 44.89 & 40.32 & 78.10 \\
 & 0SHOT-TC-Flow & 0.39 & 9.17 & 12.86 & 16.21 & \textbf{76.19} & 9.29 & 9.46 & 49.09 \\
 & 0SHOT-TC & 9.34 & 24.08 & 40.43 & 29.59 & 73.51 & 19.37 & 33.76 & 75.55 \\
 & CorEx & 41.25 & 44.27 & 44.87 & 35.61 & 32.67 & 43.58 & 31.60 & 71.87 \\
 & GLDA & 24.64 & 25.14 & 25.14 & 14.95 & 13.43 & 19.30 & 18.58 & 66.81 \\
 &  &  &  &  &  &  &  &  &  \\
\bottomrule
\end{tabular}
}
\caption{Results for classification performance. Definition of metrics can be found at Appendix \ref{appendix4}. Highest values are marked with bold font. Numbers in the brackets are $\gamma$ values (for model name) or standard deviation (for metrics). All numbers are percentages. $\gamma$ values are chosen based on the observation of the aggressiveness of the model.}
\label{classification}
\end{table*}

Table \ref{dtm} shows examples of prediction results in the format of document-topic matrix, where each value shows the probability of the topic (column) appearing in the document (row). 

\begin{table*}
\centering
\resizebox{\textwidth}{!}{
\begin{tabular}{lrrrr}
\toprule
 & ambience & service & price & food \\
\midrule
The wait staff is friendly, and the food has gotten better and better! ('food', 'service') & 0.43 & 0.69 & 0.37 & 0.73 \\
The staff is unbelievably friendly, and I dream about their Saag gosht...so good. ('service', 'food') & 0.43 & 0.71 & 0.36 & 0.51 \\
The crust is thin, the ingredients are fresh and the staff is friendly. ('food', 'service') & 0.47 & 0.69 & 0.30 & 0.68 \\
The food is outstanding and the service is quick, friendly and very professional. ('food', 'service') & 0.47 & 0.88 & 0.30 & 0.74 \\
Get the soup and a nosh (pastrami sandwich) for \$8 and you're golden. ('food', 'price') & 0.36 & 0.42 & 0.52 & 0.53 \\
Wonderful menu, warm inviting ambiance, great service the FOOD keeps me coming back! ('food', 'ambience', 'service') & 0.61 & 0.69 & 0.32 & 0.68 \\
The food was good, the service prompt, and the price very reasonable.  ('food', 'service', 'price') & 0.42 & 0.80 & 0.76 & 0.81 \\
Great food at REASONABLE prices, makes for an evening that can't be beat! ('food', 'price') & 0.47 & 0.35 & 0.78 & 0.72 \\
While the food was excellent, it wasn't cheap (though not extremely expensive either). ('food', 'price') & 0.34 & 0.32 & 0.69 & 0.75 \\
I found the food, service and value exceptional everytime I have been there. ('food', 'service', 'price') & 0.33 & 0.68 & 0.43 & 0.65 \\
\bottomrule
\end{tabular}
}
\caption{An exemplar of document-topic matrix using sentences selected from \emph{Restaurant} dataset, where rows represent documents and columns represent topics. Words in brackets after the document show the corresponding groudtruth label.}
\label{dtm}
\end{table*}

\subsection{Ablation Study and Sensitivity Analysis}
\label{as}

\begin{figure}
\centering
\includegraphics[width=\linewidth]{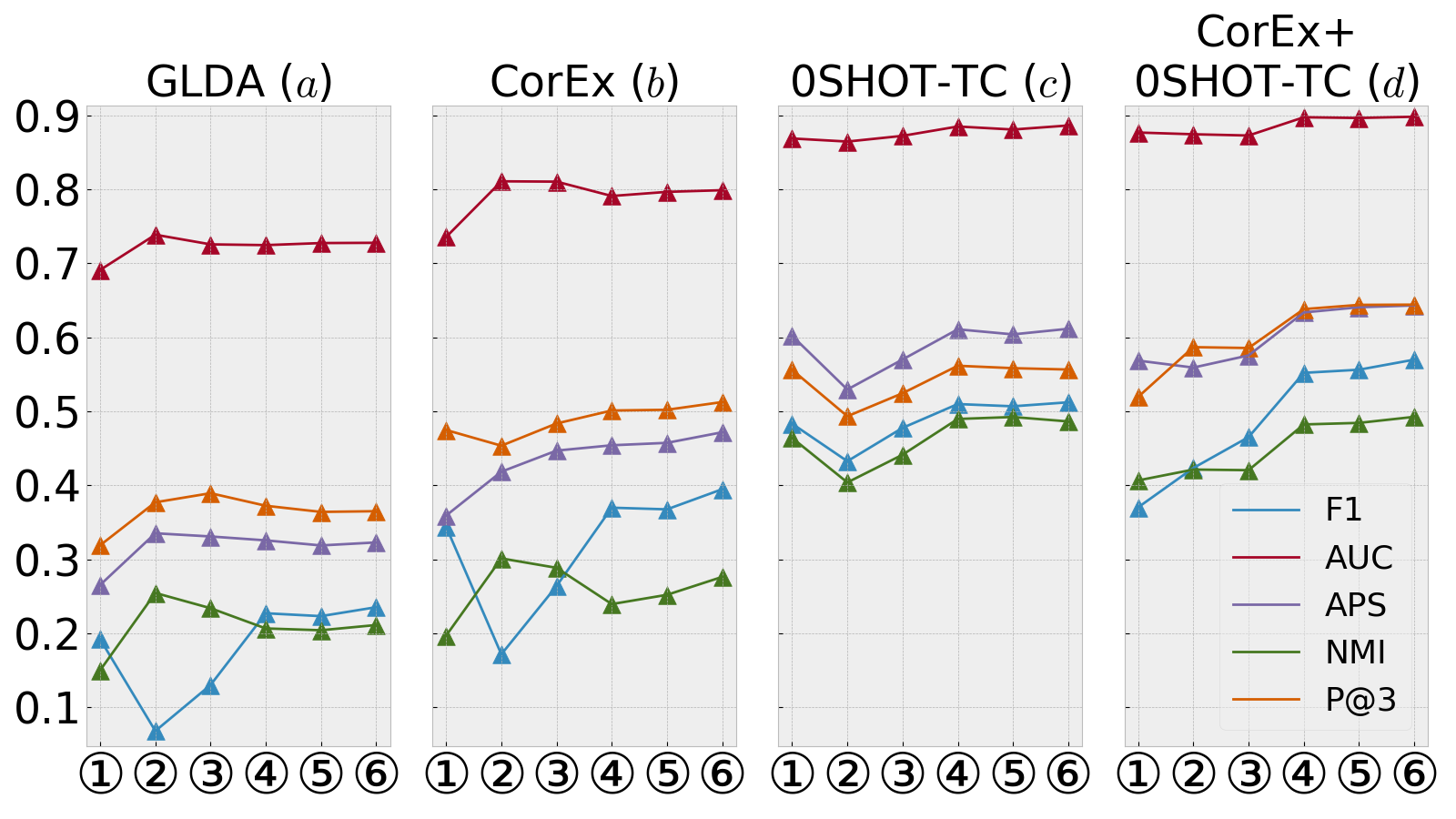}
\caption{Ablation study with respect to various backend models and components, averaged across 5 multi-label datasets excluding \emph{Heritage} dataset. Title: different backend models. X-axis: \circled{1}: Backend model only; \circled{2}: \circled{1}+BERT+encoder; \circled{3}: \circled{2}+Flow; \circled{4}: \circled{3}+VAE; \circled{5}: \circled{4}+TF-IDF; \circled{6}: \circled{5}+HPS. NMI: Normalized Mutual Information}
\label{ablation}
\end{figure}

In order to investigate further the effectiveness of each component and the sensitivity with respect to backend models, we sequentially added each component of the model and calculated its performance by averaging several key metrics across 5 multi-label (excluding \emph{Heritage}) datasets in Fig \ref{ablation}.

The results shown in Fig \ref{ablation} suggest that there is a consistent improvement of BFV models (at stage \circled{6}) with respect to various backends (at stage \circled{1}) in terms of F1, manifesting the generalizability of the benefits brought by BFV. Furthermore, after adopting the flow model and VAE in step \circled{3} and step \circled{4}, most metrics outperform that of the base models. This indicates the effectiveness of Flow-calibrated embeddings and extra two losses ($L_{R}$ and $L_{KLD}$) brought by VAE. On the other hand, the TF-IDF averaging strategy (\circled{5}) does not result in significant effects on model performance. Also, the HPS strategy (\circled{6}) does yield performance gains when GLDA and CorEx as backend models, but has limited effects on 0SHOT-TC and CorEx+0SHOT-TC backend models. 

\begin{figure}
\centering
\includegraphics[width=\linewidth]{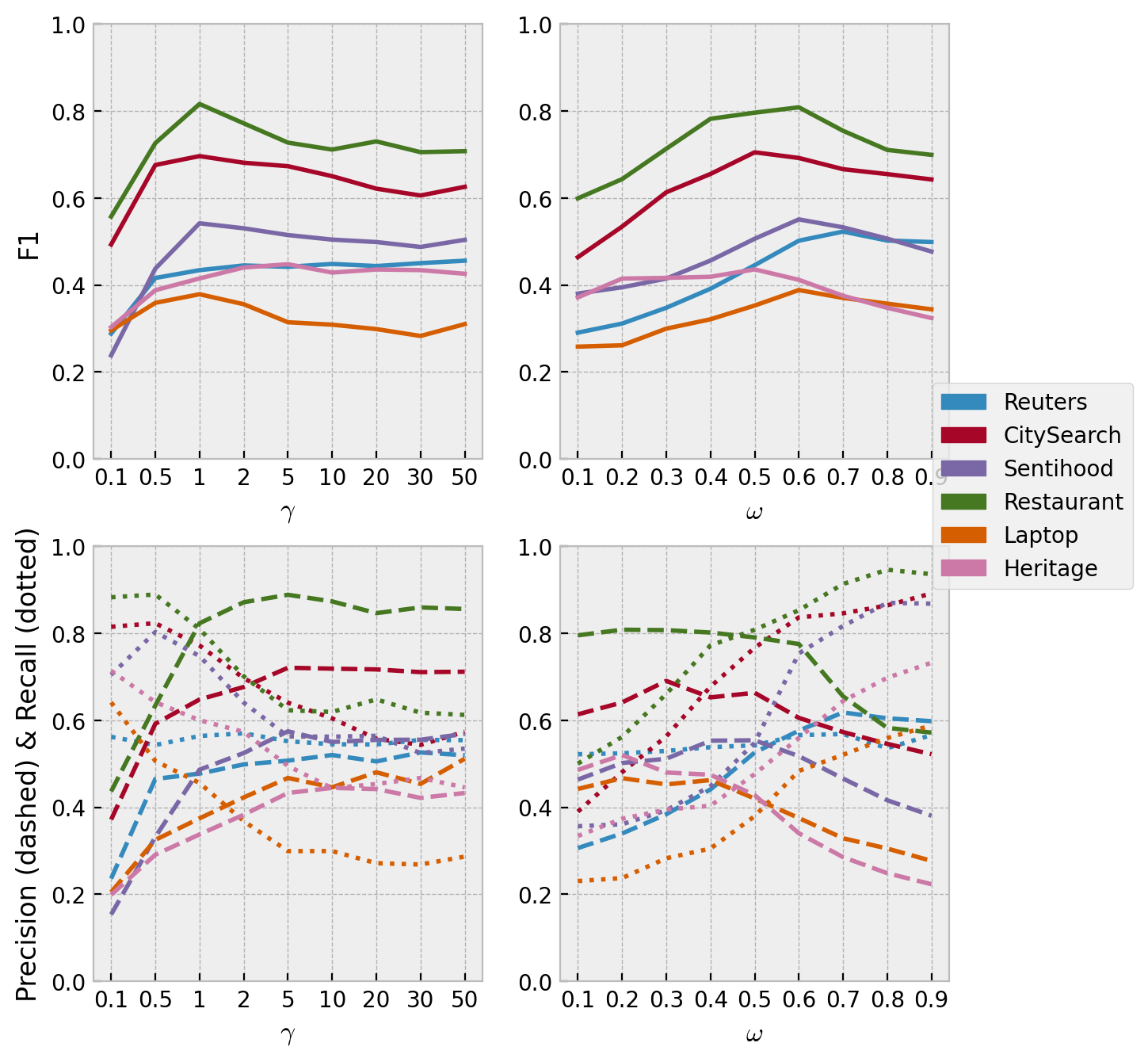}
\caption{Sensitivity with respect to $\gamma$ and $\omega$ in terms of F1 (solid line), Precision (dashed line) and Recall (dotted line) on 6 datasets}
\label{sensitivity_analysis}
\end{figure}

We also conducted a sensitivity analysis with respect to $\gamma$ and $\omega$ in Fig \ref{sensitivity_analysis}. It could be observed that the model performs best when $\gamma$ is between 0.5 and 5 and $\omega$ is between 0.5 and 0.6 for nearly all datasets. From the bottom two plots in Fig \ref{sensitivity_analysis} we can see that, $\gamma$ has a positive (negative) relationship with Precision (Recall). This relationship is reversed when it comes to $\omega$. This makes sense because $\gamma$ controls the relative weight of $L_{KLD}$ and $L_{\mathcal{T}}$ in the loss function, and $\omega$ controls the weight of the combination of two backend models. Therefore, they could influence the aggressiveness of the model. This can be used to control specificity and sensitivity of the model (with default value of $\gamma = 1$ and $\omega = 0.5$).

\subsection{Discussion}

\paragraph{Effectiveness of Flow and VAE} Based on the performance of VAE-ablated models (\textbf{BFE}, \textbf{BF} and \textbf{0SHOT-TC-Flow}) and Flow-ablated models (\textbf{BV} and \textbf{BWV}) shown in Table \ref{classification} and Figure \ref{ablation}, we can observe that, neither Flow nor VAE alone can outperform their combination (\textbf{BFV}) in terms of F1-score and APS. This reflects the importance of combining Flow and VAE to model disentangled latent variables within calibrated BERT embeddings. 

\paragraph{Mixture of Backend} Except for the benefits brought by the pre-trained language model and embeddings calibration, we also suggest that BFV can learn from both a sparser and conservative model (CorEx) and a denser and aggressive model (0SHOT-TC) simultaneously:

(1) BFV can capture complementary information from seed words in the results of CorEx in addition to the topic surface name. Specifically, Fig \ref{ablation} shows that the average of the results of CorEx and 0SHOT-TC has a lower value of F1-score (\circled{1} of $d$) compared to that of only 0SHOT-TC (\circled{1} of $c$). Note that this still holds even if we compare F1-score in \circled{3} of $d$ and \circled{1} of $c$. However, after being fully processed by BFV, the F1-score in \circled{6} of $d$ is significantly large compared to that of using only 0SHOT-TC (\circled{6} of $c$) as backend. 

\begin{figure}
\centering
\includegraphics[width=\linewidth]{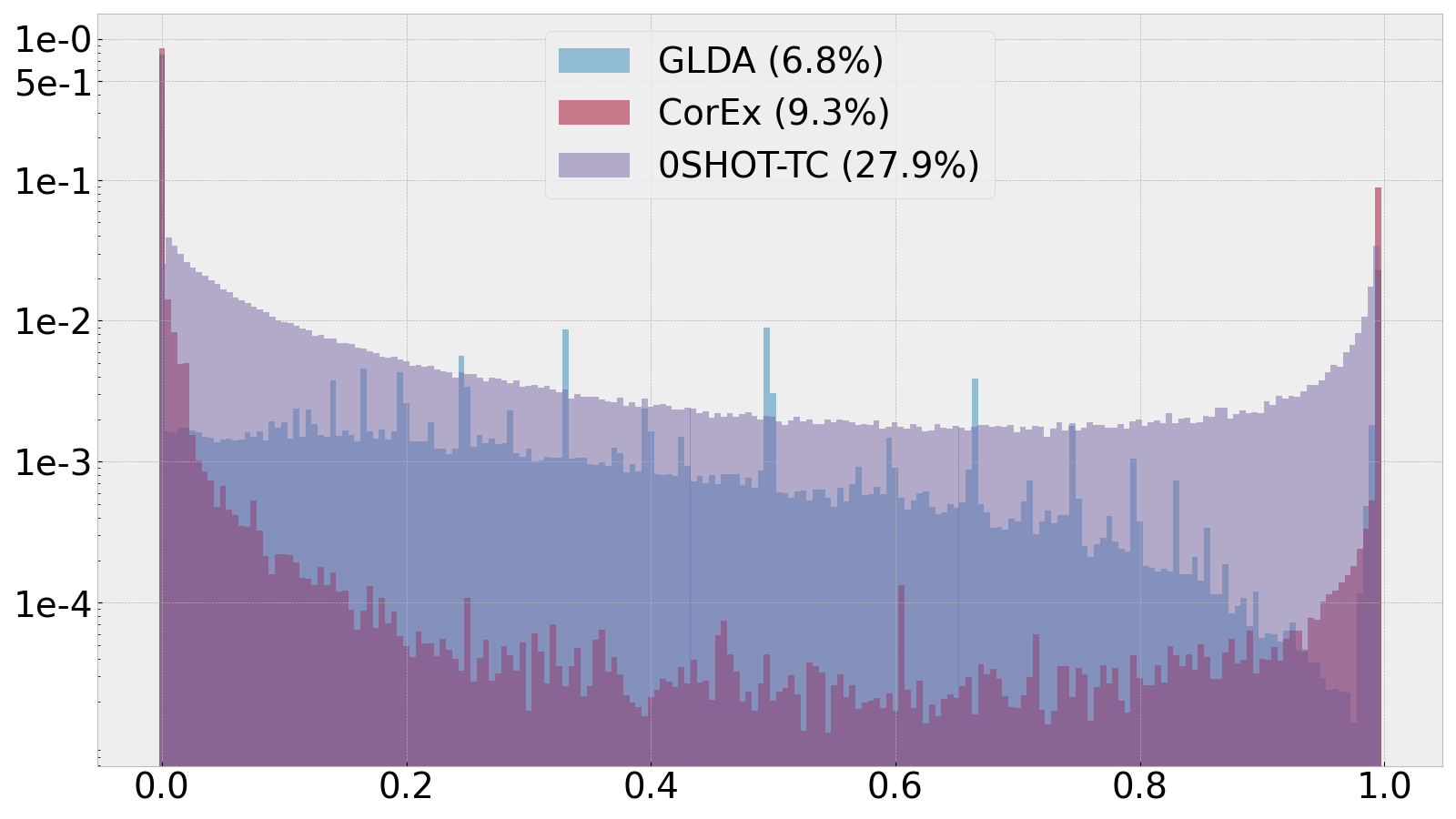}
\caption{Normalized histogram of all predicted probabilities (x-axis) by backend models. The numbers in the bracket are the percentage of positive predictions.}
\label{predprob}
\end{figure}

(2) Results from CorEx may provide regularization. We observe in Fig \ref{predprob} that, results from CorEx are sparser and conservative (most labels were predicted to be negative) than that of 0SHOT-TC. We also observe from Table \ref{classification} that CorEx has a higher Precision than GLDA (and thus GLDA is not mixed) and 0SHOT-TC in general. 


\paragraph{Different Language Models} When replacing the backend language model (distil-BERT) with BERT, RoBERTa and XLM-R, in terms of the F1-score over 10 runs averaged across the 6 datasets, we found: BERT-base (54.35\%), BERT-large (54.54\%), RoBERTa-base (55.47\%), RoBERTa-large (55.96\%) and XLM-R-base (54.50\%) compared to distil-BERT of 54.85\% (detailed dataset-specific results fan be found in Appendix \ref{appendix1}). However, we simply tested these models using the same hyper-parameters, which may cause impacts on the performance. We didn't test the effects of different combinations of language models and hyper-parameters because the potential amount of computation will be very large for grid search. Thus, it is difficult to draw conclusions at this stage and we propose to leave in-depth analysis as future work where different characteristics of the backend language model should be thoroughly considered.

\section{Conclusion}



We presented BFV, a WSMLTC model, that uses a VAE framework to reconstruct BERT-produced and flow-calibrated sentence embeddings under the guidance of the averaged results of CorEx and 0SHOT-TC. It can significantly outperform other WSMLTC models in key metrics and achieve a approximately 84\% performance of a fully-supervised model in terms of macro F1-score evaluated on 6 datasets. We found the improvements are mainly due to: (1) combining BERT and VAE framework, (2) mapping the sentence embeddings into a standard Gaussian space to better fit the overall objective of the VAE framework and (3) learning simultaneously from the results produced by a mixture of backends. As the input of surface name of topics and seed words are only used to produce $\mathcal{T}$, BFV can be viewed as a post-processing model to refine an already made document-topic matrix.

One limitation of the current BFV model is that it does not explicitly model the relationship between topics and words. Thus, relevant tasks such as calculating topic coherence and selecting keywords for each topic cannot be done directly. Another drawback of the BFV model is that it does not take into account the dependencies and hierarchies within topics. This may limit the model's performance for datasets in which labels are correlated. 


In the future, BFV could be improved by reconstructing embeddings of each words in a sentence rather than an embedding of a sentence as a whole. This can also potentially expand this method into an applicable generative model for real sentence generation. Further work could also evaluate the performance of the model using embeddings made by language models other than distil-BERT.

\section*{Acknowledgements}
We would like to thank the reviewers for their constructive feedback. We would also want to thank Dr Pontus Stenetorp for his helpful advice.
\newpage
\bibliography{anthology,custom}
\bibliographystyle{acl_natbib}

\newpage

\section*{Appendix}
\appendix
\section{Variational AutoEncoder (VAE)}
\label{appendix5}

Variational AutoEncoder (VAE) \cite{kingma2013auto} has been widely used as an unsupervised generative method, especially in computer vision filed. It has been shown that $\beta$-VAE \cite{higgins2016beta} and its variants are capable of finding visual disentangled latent representations which remain invariant to some transformations \cite{burgess2018understanding}. This section provides a brief review of VAE and $\beta$-VAE.


VAE starts by solving the evidence (or marginal likelihood) $p_\theta(x)$ in Variational Inference (VI) problems involving latent variable $Z$. In particular, the evidence $p_\theta(x)=\int p(x,z)dz$ is often intractable to compute in practice due to computational cost. In VI, $p_\theta(x)$ can be approximated by introducing variational distributions and building a lower bound $\mathcal{L}$ to reframe the problem as in Eq. \ref{eq1} based on Jensen's inequality.

\begin{align}
\label{eq1}
\begin{split}
    logp_\theta(x) & = log(\mathbb{E}_{q_{\phi}(z|x)}[p_\theta(x|z)\frac{p(z)}{q_{\phi}(z|x)}])\\
                   & \geq  \mathbb{E}_{q_{\phi}}[logp_\theta(x|z)]-KLD(q_{\phi}(z|x)||p(z))\\
                   & = \mathcal{L}(\theta,\phi;x)\\ & = -(L_{R}+L_{KLD})
\end{split}
\end{align}

$L_{R}$ and $L_{KLD}$ refer to reconstruction loss and KLD loss respectively. By considering $q_{\phi}(z|x)$ as a Gaussian probabilistic encoder, $p(z)$ as a standard Gaussian prior $\mathcal{N}(0,I)$ and $p_{\theta}(x|z)$ as a probabilistic decoder, the objective can be defined as $\underset{\theta,\phi}{\max} \  \mathcal{L}(\theta,\phi;x)$. In particular, this objective function aims at finding optimal parameters $\theta$ and $\phi$ to maximize the lower bound $\mathcal{L}$ which in turn approximates the log-probability of the data $logp_\theta(x)$. In addition, in order to estimate the gradients of the lower bound with respect to $\phi$ more smoothly, a parameterization trick is applied: $z_{i} \sim q_{\phi}(z_{i}|x) = \mu_{i}+\sigma_{i}\epsilon_{i}$, where $\epsilon_{i} \sim \mathcal{N}(0,1)$.

In $\beta$-VAE, a hyper-parameter $\beta$ is added into the objective function as shown in Eq. \ref{eq3}. Usually $\beta > 1$ will result in more disentangled representations and when $\beta = 1$, the $\beta$-VAE is equivalent to the vanilla VAE model \cite{burgess2018understanding}.

\begin{align}
\label{eq3}
    \mathcal{L}(\theta,\phi;x,\beta) = -(L_{R} + \beta L_{KLD})
\end{align}

\section{Comparison with Weakly-supervised Multi-Class Methods}
\label{appendix2}

Here we compare our model with the following strong weakly-supervised multi-class models on MLTC datasets:

WeSTClass \cite{meng2018weakly}: a WSTC model that has been briefly introduced in the related-work section. It can receive inputs of topic surface name, keywords or limited amount of documents. We used the keywords as input to it. We replaced its last softmax layer with sigmoid layer and used the same threshold (0.5).

X-Class \cite{wang2020x}: this methods is based on aligning document representation and class representation. It uses class surface names as input and can generage pseudo labels to train a text classifier. We used its X-Class-Align version which uses a Gaussian Mixture Model (GMM) to make final predictions. We modified its final to output unnormalized probabilities and used the threshold of 0.5.

LOTClass \cite{meng2020text}: it is based on Masked Category Prediction (MCP) task and a subsequent self-training to perform WSTC with only label surface names. We replaced its last softmax layer with sigmoid and used the same threshold of 0.5. We also tested transforming it into a binary relevance task by only using one label surface name each time, which results in similar performance and therefore is not reported. 

We used default hyper-parameters for all three models. We slightly modified keywords and topic surface names if the model has a different specification, otherwise we used the same keywords and topic surface names as used in our model. The results of comparison is presented in Table \ref{compare}. It can be shown that there is a large performance margin between our model and the weakly-supervised multi-class models on multi-label tasks in terms of key metrics such as F1-score.

\begin{table*}[ht]
\centering
\resizebox{\textwidth}{!}{
\begin{tabular}{llllllll}
\toprule
{} & {} & {ACC} & {F1} & {Recall} & {Precision} & {APS} & {AUC} \\
\midrule
\multirow[c]{5}{*}{Reuters} & BFV (20) & \textbf{44.65} & \textbf{44.99} & \textbf{55.38} & \textbf{51.58} & \textbf{56.18} & \textbf{95.86} \\
 & X-Class & 27.63 & 33.83 & 38.05 & 40.64 & 30.44 & 67.84 \\
 & WeSTClass & 6.16 & 1.12 & 2.60 & 1.84 & 2.99 & 48.18 \\
 & LOTClass & 0.50 & 0.06 & 2.52 & 0.07 & 7.15 & 68.46 \\
 &  &                                                                                                      &                                                                                                      &                                                                                                      &                                                                                                      &                                                                                                      &                                                                                                      \\
\multirow[c]{5}{*}{CitySearch} & BFV (1) & \textbf{61.07} & \textbf{69.48} & \textbf{76.50} & \textbf{64.92} & \textbf{77.25} & \textbf{92.37} \\
 & X-Class & 49.77 & 16.62 & 25.00 & 12.44 & 18.78 & 50.00 \\
 & WeSTClass & 25.49 & 22.26 & 23.31 & 24.21 & 18.99 & 47.50 \\
 & LOTClass & 33.33 & 25.43 & 36.17 & 32.93 & 36.23 & 69.49 \\
 &  &                                                                                                      &                                                                                                      &                                                                                                      &                                                                                                      &                                                                                                      &                                                                                                      \\
\multirow[c]{5}{*}{Sentihood} & BFV (1) & 42.94 & \textbf{53.54} & \textbf{75.58} & \textbf{47.21} & \textbf{59.10} & \textbf{92.12} \\
 & X-Class & \textbf{55.94} & 17.29 & 16.29 & 46.58 & 12.48 & 54.32 \\
 & WeSTClass & 8.92 & 6.50 & 12.30 & 10.61 & 7.34 & 51.28 \\
 & LOTClass & 10.93 & 4.26 & 11.72 & 4.98 & 10.05 & 58.75 \\
 &  &                                                                                                      &                                                                                                      &                                                                                                      &                                                                                                      &                                                                                                      &                                                                                                      \\
\multirow[c]{5}{*}{Restaurant} & BFV (1) & \textbf{68.85} & \textbf{80.49} & \textbf{80.58} & \textbf{80.88} & \textbf{89.73} & \textbf{95.21} \\
 & X-Class & 38.75 & 13.96 & 24.92 & 9.70 & 24.75 & 50.03 \\
 & WeSTClass & 26.38 & 23.64 & 24.94 & 25.91 & 25.90 & 49.72 \\
 & LOTClass & 35.25 & 13.53 & 25.16 & 37.41 & 44.58 & 69.51 \\
 &  &                                                                                                      &                                                                                                      &                                                                                                      &                                                                                                      &                                                                                                      &                                                                                                      \\
\multirow[c]{5}{*}{Laptop} & BFV (2) & 45.72 & \textbf{36.40} & \textbf{38.59} & \textbf{42.75} & \textbf{39.25} & \textbf{73.38} \\
 & X-Class & \textbf{68.07} & 16.20 & 20.00 & 13.61 & 11.88 & 50.00 \\
 & WeSTClass & 22.65 & 15.01 & 18.93 & 19.53 & 12.73 & 49.27 \\
 & LOTClass & 9.03 & 10.41 & 28.27 & 18.04 & 15.83 & 55.45 \\
 &  &                                                                                                      &                                                                                                      &                                                                                                      &                                                                                                      &                                                                                                      &                                                                                                      \\
\multirow[c]{5}{*}{Heritage} & BFV (5) & 40.10 & \textbf{44.20} & \textbf{50.34} & \textbf{41.75} & \textbf{46.68} & \textbf{81.55} \\
 & X-Class & \textbf{41.25} & 7.31 & 12.50 & 5.16 & 10.01 & 50.22 \\
 & WeSTClass & 10.12 & 8.47 & 9.59 & 12.07 & 9.80 & 48.71 \\
 & LOTClass & 10.25 & 5.48 & 15.42 & 4.15 & 16.41 & 61.87 \\
 &  &                                                                                                      &                                                                                                      &                                                                                                      &                                                                                                      &                                                                                                      &                                                                                                      \\
\bottomrule
\end{tabular}
}
\caption{Performance of BFV in comparison with weakly-supervised multi-class models on multi-label datasets. All numbers are percentages.}
\label{compare}
\end{table*}

\section{Further Evaluation}
\label{appendix1}

\begin{figure*}[ht]
\includegraphics[width=\linewidth]{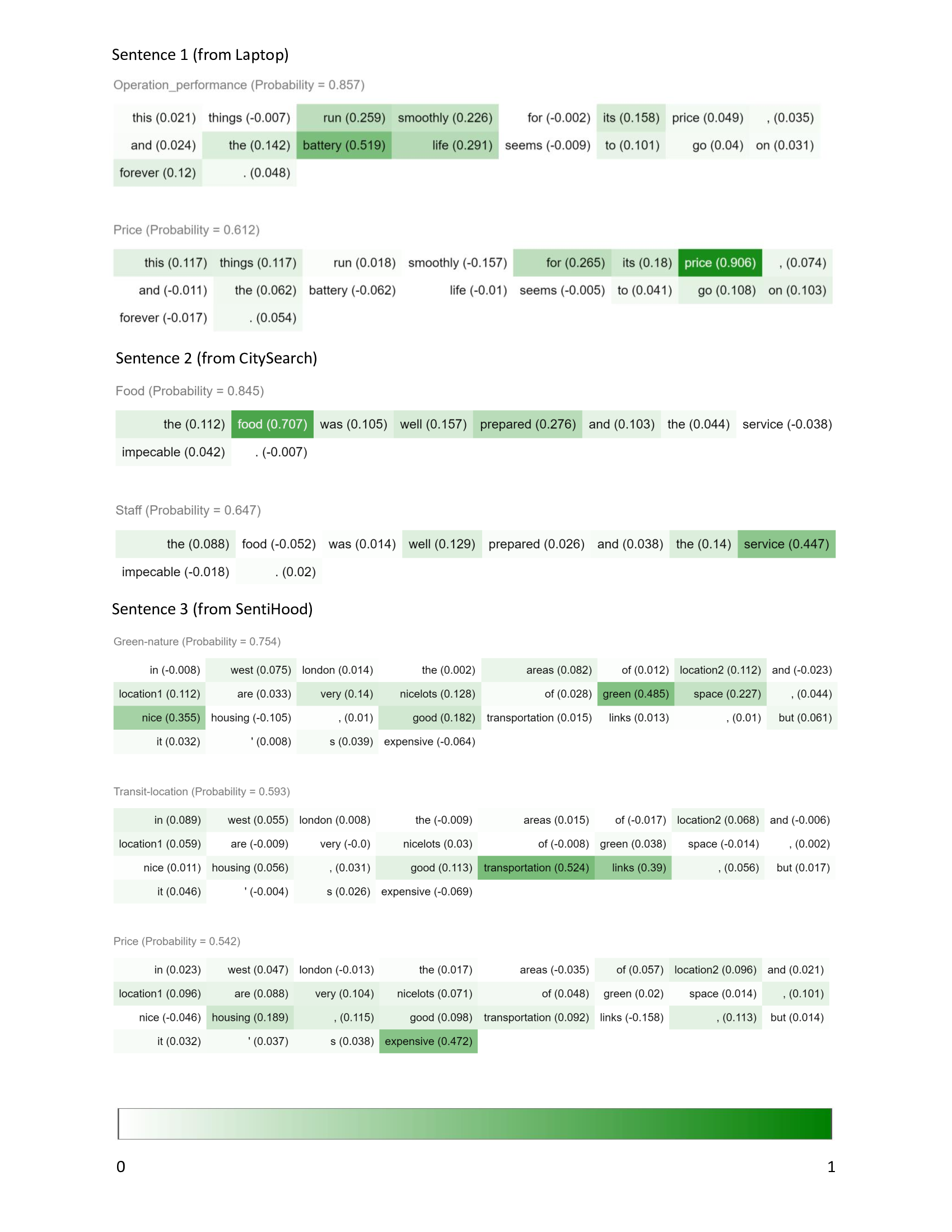}
\caption{Qualitative analysis of the correct predictions made by BFV model using Integrated Gradients (IG). Numbers with bracket attached to each words are gradients given by the IG}
\label{qualitative_analysis}
\end{figure*}

\paragraph{Qualitative Analysis} To further check the validity of the BFV model, we used Integrated Gradients (IG) \cite{sundararajan2017axiomatic} to calculate the attributions of each word with respect to predicted topics. In Fig \ref{qualitative_analysis}, we applied this inspection on 3 sentences from different datasets as examples. From Fig \ref{qualitative_analysis}, it can be seen that the gradients of words can largely align with human intuition.

\paragraph{Clustering Results} We also tested our model's performance in terms of clustering metrics. In calculating clustering metrics, we only used samples with single label in the whole dataset. Table \ref{clustering} shows the models' performance measured by clustering metrics.

\begin{table*}[ht]
\centering
\resizebox{0.7\textwidth}{!}{
\begin{tabular}{lllllll}
\toprule
{} & {} & {Homogeneity} & {Completeness} & {NMI} & {Adj MI} & {Adj Rand} \\
\midrule
\multirow[c]{12}{*}{Reuters} & BFV (20) & 67.72 (0.39) & 58.73 (0.74) & 62.90 (0.56) & 61.87 (0.58) & 69.99 (1.86)\\
 & BWV (20) & 66.08 (0.38) & 54.94 (0.62) & 60.00 (0.52) & 58.87 (0.53) & 64.90 (1.45)\\
 & BV (20) & \textbf{67.87} (0.38) & \textbf{61.73} (0.64) & \textbf{64.66} (0.48) & \textbf{63.71} (0.5) & \textbf{73.34} (0.83)\\
 & BFVCLS (20) & 64.17 (1.1) & 60.21 (1.98) & 62.11 (1.41) & 61.23 (1.43) & 71.00 (3.85)\\
 & BFE (20) & 63.29 (0.26) & 56.39 (0.33) & 59.64 (0.28) & 58.56 (0.29) & 69.03 (0.6)\\
 & BF & 22.45 & 17.47 & 19.65 & 17.93 & 9.12 \\
 & 0SHOT-TC+CorEx & 58.74 & 45.12 & 51.04 & 49.62 & 54.98 \\
 & 0SHOT-TC-Flow & 11.36 & 7.22 & 8.83 & 6.33 & 3.99 \\
 & 0SHOT-TC & 63.93 & 52.93 & 57.91 & 56.78 & 64.09 \\
 & CorEx & 34.40 & 25.24 & 29.12 & 27.23 & 22.75 \\
 & GLDA & 46.96 & 31.69 & 37.85 & 35.98 & 23.46 \\
 &  &  &  &  &  &  \\
\multirow[c]{12}{*}{CitySearch} & BFV (1) & \textbf{51.71} (1.19) & \textbf{48.99} (1.17) & \textbf{50.31} (1.16) & \textbf{50.21} (1.16) & \textbf{60.21} (1.37)\\
 & BWV (1) & 44.91 (1.24) & 42.36 (1.64) & 43.60 (1.44) & 43.48 (1.44) & 51.63 (1.94)\\
 & BV (1) & 43.50 (1.7) & 43.19 (2.29) & 43.32 (1.7) & 43.20 (1.7) & 51.72 (2.89)\\
 & BFVCLS (1) & 40.11 (1.95) & 44.07 (1.8) & 41.98 (1.68) & 41.85 (1.68) & 50.32 (2.88)\\
 & BFE (1) & 41.06 (0.81) & 42.58 (0.87) & 41.81 (0.81) & 41.67 (0.81) & 50.94 (0.78)\\
 & BF & 0.69 & 2.77 & 1.11 & 0.73 & 2.59 \\
 & 0SHOT-TC+CorEx & 35.86 & 33.45 & 34.61 & 34.47 & 40.80 \\
 & 0SHOT-TC-Flow & 14.94 & 13.61 & 14.25 & 14.06 & 13.96 \\
 & 0SHOT-TC & 43.24 & 46.24 & 44.69 & 44.56 & 54.24 \\
 & CorEx & 15.92 & 14.07 & 14.94 & 14.76 & 8.28 \\
 & GLDA & 7.36 & 6.24 & 6.75 & 6.56 & 6.75 \\
 &  &  &  &  &  &  \\
\multirow[c]{12}{*}{Sentihood} & BFV (1) & 53.83 (1.48) & 51.72 (1.48) & 52.76 (1.46) & 52.34 (1.47) & 49.13 (2.69)\\
 & BWV (1) & 48.83 (1.39) & 48.41 (1.15) & 48.62 (1.25) & 48.15 (1.26) & 37.53 (3.33)\\
 & BV (1) & 45.42 (3.17) & 48.56 (2.38) & 46.92 (2.61) & 46.41 (2.64) & 28.08 (5.67)\\
 & BFVCLS (1) & 34.68 (2.8) & 41.01 (2.71) & 37.57 (2.69) & 36.96 (2.73) & 20.68 (5.11)\\
 & BFE (1) & 42.67 (0.75) & 44.54 (0.76) & 43.59 (0.75) & 43.06 (0.76) & 28.01 (0.8)\\
 & BF & 3.37 & 5.26 & 4.11 & 3.44 & 3.03 \\
 & 0SHOT-TC+CorEx & 52.08 & 50.04 & 51.04 & 50.60 & 49.09 \\
 & 0SHOT-TC-Flow & 5.42 & 5.56 & 5.49 & 4.61 & 3.59 \\
 & 0SHOT-TC & \textbf{57.08} & \textbf{56.77} & \textbf{56.92} & \textbf{56.53} & \textbf{49.87} \\
 & CorEx & 24.57 & 25.50 & 25.03 & 24.33 & 11.79 \\
 & GLDA & 18.67 & 16.53 & 17.53 & 16.84 & 15.11 \\
 &  &  &  &  &  &  \\
\multirow[c]{12}{*}{Restaurant} & BFV (1) & \textbf{59.23} (1.29) & \textbf{57.23} (1.85) & \textbf{58.21} (1.5) & \textbf{58.13} (1.51) & \textbf{68.13} (1.84)\\
 & BWV (1) & 50.04 (1.78) & 44.93 (1.84) & 47.34 (1.81) & 47.25 (1.81) & 53.10 (2.78)\\
 & BV (1) & 46.07 (1.34) & 44.48 (1.85) & 45.25 (1.5) & 45.15 (1.5) & 49.78 (4.81)\\
 & BFVCLS (1) & 41.03 (2.32) & 44.44 (2.63) & 42.63 (2.08) & 42.51 (2.08) & 50.98 (4.88)\\
 & BFE (1) & 45.86 (0.79) & 44.64 (0.84) & 45.24 (0.81) & 45.14 (0.81) & 52.47 (1.19)\\
 & BF & 1.41 & 27.83 & 2.69 & 2.45 & 1.50 \\
 & 0SHOT-TC+CorEx & 49.44 & 45.98 & 47.65 & 47.55 & 56.46 \\
 & 0SHOT-TC-Flow & 1.79 & 1.59 & 1.68 & 1.50 & 3.32 \\
 & 0SHOT-TC & 54.94 & 56.52 & 55.72 & 55.63 & 67.55 \\
 & CorEx & 19.13 & 18.50 & 18.81 & 18.66 & 5.37 \\
 & GLDA & 9.93 & 8.21 & 8.99 & 8.83 & 7.97 \\
 &  &  &  &  &  &  \\
\multirow[c]{12}{*}{Laptop} & BFV (2) & \textbf{22.28} (0.95) & 21.90 (1.11) & \textbf{22.09} (1.02) & \textbf{21.78} (1.02) & \textbf{14.27} (0.77)\\
 & BWV (2) & 20.38 (0.58) & 21.46 (0.51) & 20.90 (0.54) & 20.58 (0.55) & 12.78 (1.18)\\
 & BV (2) & 20.45 (0.99) & 22.04 (0.98) & 21.20 (0.84) & 20.87 (0.84) & 12.75 (1.75)\\
 & BFVCLS (2) & 12.52 (1.12) & 17.58 (1.59) & 14.60 (1.09) & 14.25 (1.11) & 6.13 (0.95)\\
 & BFE (2) & 18.67 (0.69) & 21.48 (0.77) & 19.98 (0.73) & 19.63 (0.73) & 10.42 (0.43)\\
 & BF & 2.75 & \textbf{44.62} & 5.18 & 4.85 & 1.46 \\
 & 0SHOT-TC+CorEx & 18.45 & 19.57 & 19.00 & 18.66 & 11.48 \\
 & 0SHOT-TC-Flow & 0.58 & 0.62 & 0.60 & 0.18 & -0.25 \\
 & 0SHOT-TC & 15.33 & 18.53 & 16.78 & 16.41 & 8.40 \\
 & CorEx & 9.85 & 11.51 & 10.61 & 10.23 & 3.80 \\
 & GLDA & 4.11 & 3.86 & 3.98 & 3.61 & 2.24 \\
 &  &  &  &  &  &  \\
\multirow[c]{12}{*}{Heritage} & BFV (5) & 26.95 (0.7) & 26.49 (0.77) & 26.72 (0.73) & 26.31 (0.74) & 23.98 (1.29)\\
 & BWV (5) & 25.24 (0.38) & 24.61 (0.35) & 24.92 (0.36) & 24.50 (0.37) & 22.20 (0.5)\\
 & BV (5) & \textbf{28.99} (0.74) & \textbf{28.82} (0.71) & \textbf{28.90} (0.7) & \textbf{28.50} (0.7) & \textbf{25.30} (1.66)\\
 & BFVCLS (5) & 26.43 (0.89) & 27.59 (0.96) & 26.99 (0.84) & 26.57 (0.84) & 22.10 (2.16)\\
 & BFE (5) & 26.92 (0.58) & 27.04 (0.6) & 26.98 (0.59) & 26.57 (0.59) & 23.28 (0.61)\\
 & BF & 0.69 & 3.40 & 1.15 & 0.65 & -0.17 \\
 & 0SHOT-TC+CorEx & 21.22 & 20.78 & 21.00 & 20.56 & 17.60 \\
 & 0SHOT-TC-Flow & 1.08 & 1.63 & 1.30 & 0.61 & -0.18 \\
 & 0SHOT-TC & 20.70 & 20.69 & 20.70 & 20.25 & 16.25 \\
 & CorEx & 15.07 & 15.97 & 15.50 & 15.01 & 8.00 \\
 & GLDA & 6.62 & 6.41 & 6.51 & 5.99 & 4.52 \\
 &  &  &  &  &  &  \\
\bottomrule
\end{tabular}
}
\caption{Results for clustering performance. Highest values are marked with bold font. Numbers in the brackets are $\gamma$ values (in column) or standard deviation (in matrix). All numbers are percentages.}
\label{clustering}
\end{table*}

\begin{figure*}[ht]
\includegraphics[width=\linewidth]{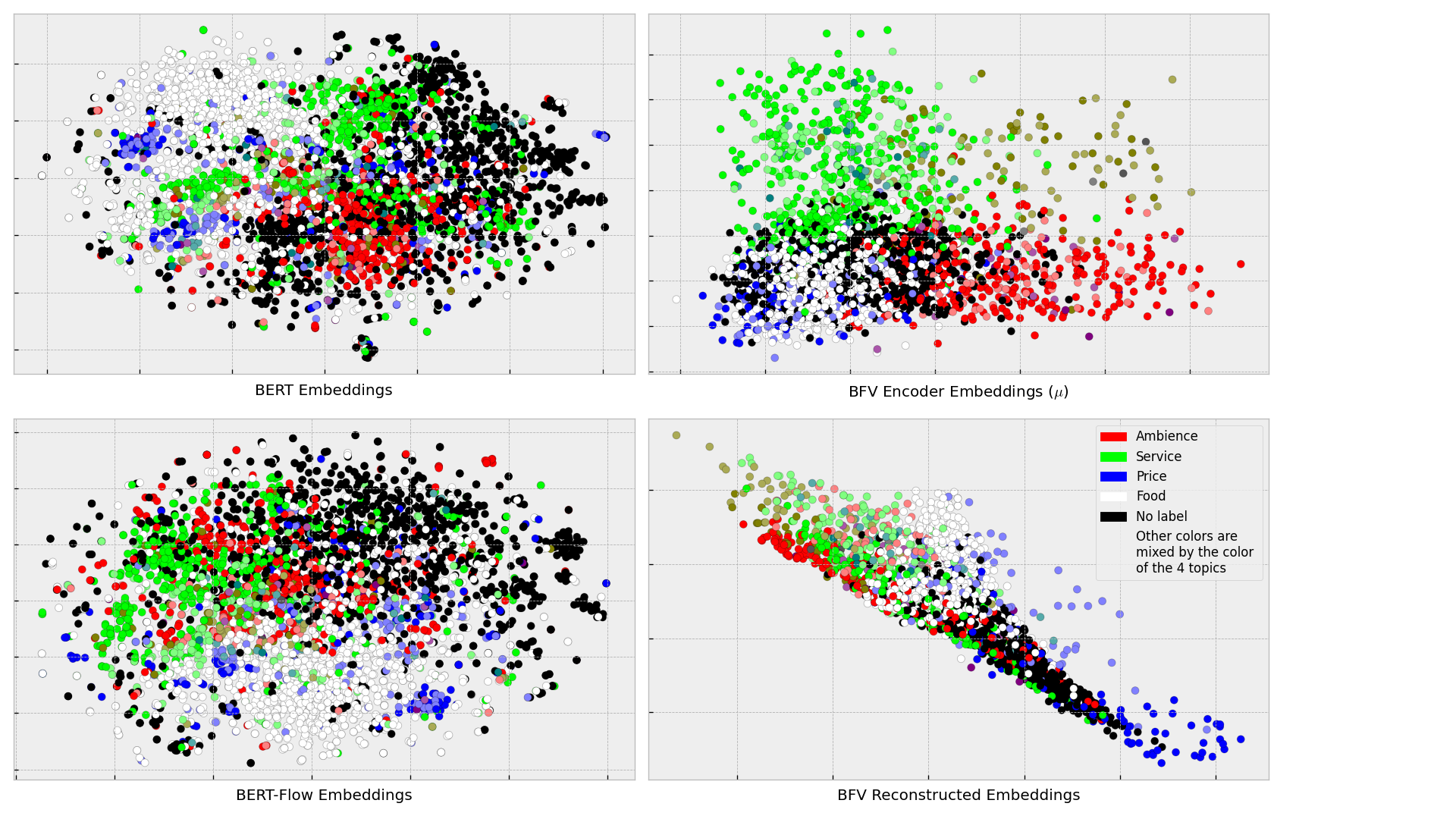}
\caption{Visualization of embeddings produced at different stages of the model by T-SNE using the \emph{Restaurant} dataset. Red, Green, Blue and White are used to represent topics 'ambience', 'service', 'price' and 'food' in the dataset. Colors other than Red, Blue, Green and White are mixed by the color of the corresponding topics when there are more than one topics assigned to the document.}
\label{tsne}
\end{figure*}

\paragraph{Embeddings Visualization} Embeddings produced at different stages of the model are visualized by T-SNE in Figure \ref{tsne}, where the shift of the embeddings during the process could be observed.

\paragraph{Other Language Models} We tested the model's performance with different backend language models. Table \ref{olm} shows the model's performance on different datasets in terms of F1-score when using other language models. However, we simply tested these models using the same hyper-parameters, which may cause impacts on the performance.

\begin{table*}
\begin{tabular}{lrrrrr}
\toprule
 & BERT-base & BERT-large & RoBERTa-base & RoBERTa-large & XLM-R-base \\
\midrule
Reuters & 42.12 & 43.53 & 40.22 & 44.16 & 42.16 \\
CitySearch & 69.52 & 69.05 & 70.96 & 71.73 & 70.42 \\
Sentihood & 54.49 & 52.14 & 55.34 & 54.48 & 53.54 \\
Restaurant & 79.02 & 80.72 & 81.14 & 83.53 & 79.29 \\
Laptop & 38.49 & 38.40 & 40.56 & 39.45 & 39.16 \\
Heritage & 42.45 & 43.41 & 44.57 & 42.41 & 42.43 \\
\bottomrule
\end{tabular}
\caption{Models performance on 6 datasets with different backend language models in terms of F1-score. All numbers are percentages}
\label{olm}
\end{table*}

\section{Supplementary Information of the Datasets}
\label{appendix3}

\begin{figure*}[ht]
\includegraphics[width=\linewidth]{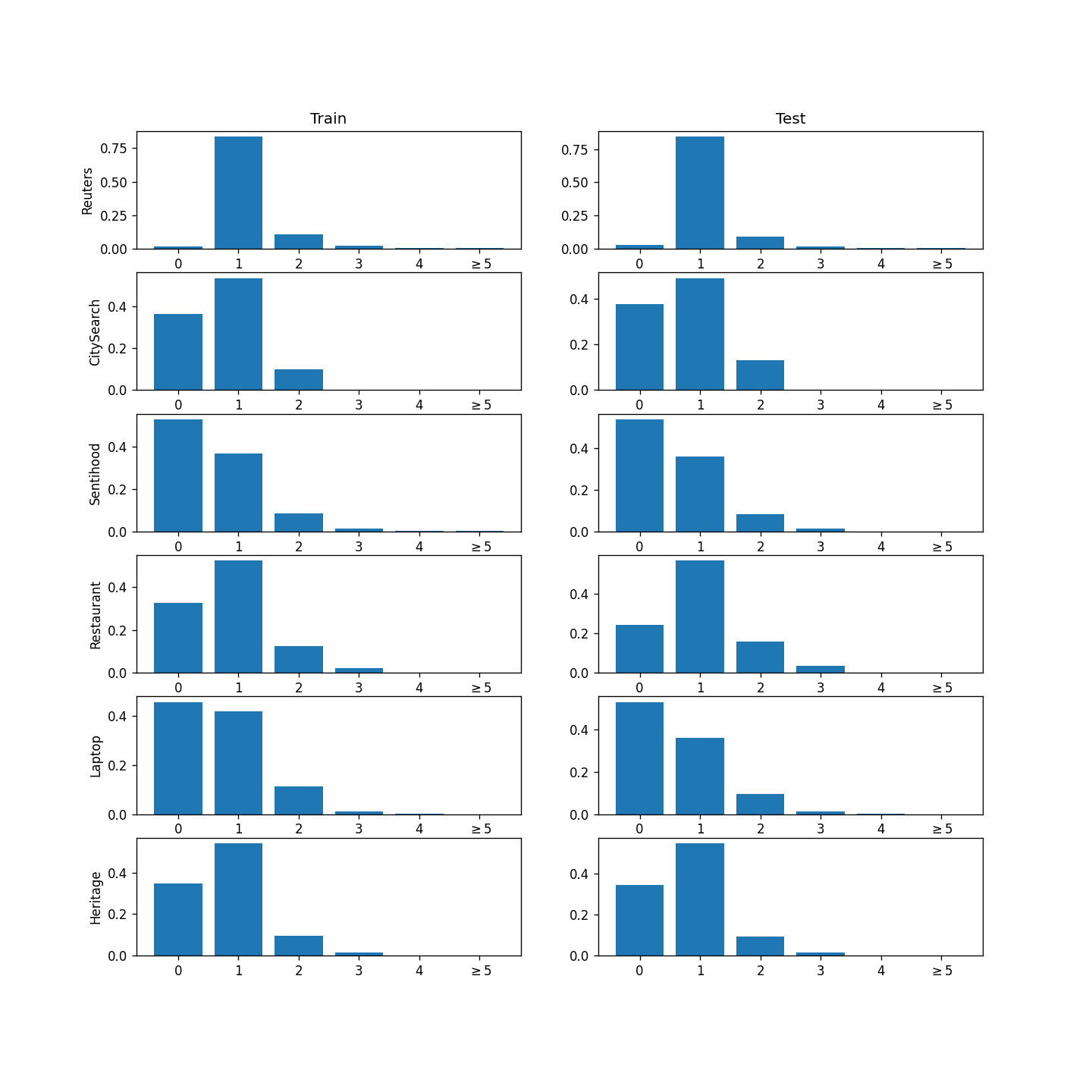}
\caption{Distribution of the amount of topics in the samples in percentage.}
\label{mlrd}
\end{figure*}

Fig \ref{mlrd} displays the distribution of the amount of topics in the samples in percentage for different datasets used in this paper.

In regard to the \emph{Heritage} dataset, all annotators (authors of the paper) have checked the correctness of labelling and are fully aware of the risks to participant.

\section{Definition of Evaluation Metrics}
\label{appendix4}

In evaluating our model in Table \ref{classification}, we have the following metrics. Assuming $y_{i} \in \mathbb{R}^{C}$ is the ground truth label and $\hat{y_{i}}$ is the prediction for $ith$ sample, where $C$ is the number of classes:

\textbf{ACC}:
\begin{equation*}
\text{Accuracy} = \frac{1}{n} \sum_{i=1}^{n} I(y_{i} = \hat{y_{i}})
\end{equation*}

\textbf{HS}:
\begin{equation*}
\text{Hamming Score} = \frac{1}{n} \sum_{i=1}^{n} \frac{\lvert y_{i} \cap \hat{y_{i}}\rvert}{\lvert y_{i} \cup \hat{y_{i}}\rvert}
\end{equation*}

\textbf{Precision}:
\begin{equation*}
\text{Precision} = \frac{1}{n} \sum_{i=1}^{n} \frac{\lvert y_{i} \cap \hat{y_{i}}\rvert}{\lvert \hat{y_{i}}\rvert}
\end{equation*}

\textbf{Recall}:
\begin{equation*}
\text{Recall} = \frac{1}{n} \sum_{i=1}^{n} \frac{\lvert y_{i} \cap \hat{y_{i}}\rvert}{\lvert y_{i}\rvert}
\end{equation*}

\textbf{F1-score}:
\begin{equation*}
F_{1} = \frac{1}{n} \sum_{i=1}^{n} \frac{2 \lvert y_{i} \cap \hat{y_{i}}\rvert}{\lvert y_{i}\rvert + \lvert \hat{y_{i}}\rvert}
\end{equation*}

\textbf{APS}:
\begin{equation*}
\text{Average Precision Score}\footnote{\url{https://scikit-learn.org/stable/modules/generated/sklearn.metrics.average_precision_score.html}}
\end{equation*}

\textbf{AUC}:
\begin{equation*}
\text{ROC-AUC score}\footnote{\url{https://scikit-learn.org/stable/modules/generated/sklearn.metrics.roc_auc_score.html}}
\end{equation*}

\textbf{P@3}:
\begin{equation*}
\text{Mean Average Precision @ k = 3}\footnote{\url{https://github.com/benhamner/Metrics/blob/master/Python/ml_metrics/average_precision.py}}
\end{equation*}

\textbf{Homogeneity}:

\begin{equation*}
\text{homogeneity score}\footnote{\url{https://scikit-learn.org/stable/modules/generated/sklearn.metrics.homogeneity_score.html\#sklearn.metrics.homogeneity_score}}
\end{equation*}

\textbf{Completeness}:

\begin{equation*}
\text{completeness score}\footnote{\url{https://scikit-learn.org/stable/modules/generated/sklearn.metrics.completeness_score.html\#sklearn.metrics.completeness_score}}
\end{equation*}

\textbf{NMI}:

\begin{equation*}
v = \frac{2 \times \text{homogeneity} \times \text{completeness}}{\text{homogeneity} + \text{completeness}}
\end{equation*}

\textbf{Adj MI}:

\begin{equation*}
\text{Adjusted Mutual Info Score}\footnote{\url{https://scikit-learn.org/stable/modules/generated/sklearn.metrics.adjusted_mutual_info_score.html}}
\end{equation*}

\textbf{Adj Rand}:
\begin{equation*}
\text{Adjusted Rand Score}\footnote{\url{https://scikit-learn.org/stable/modules/generated/sklearn.metrics.adjusted_rand_score.html}}
\end{equation*}
\end{document}